\documentclass[runningheads]{llncs}
\usepackage{graphicx}
\usepackage{multirow}
\usepackage{footnote}
\usepackage{footmisc}
\usepackage{cite}
\usepackage[font=small,skip=1pt]{caption}
\usepackage{booktabs}
\usepackage{lipsum}

\usepackage{subfig}

\begin{document}
\title{Integrating Crowdsourcing and Active Learning for Classification of Work-Life Events from Tweets}
\titlerunning{Crowdsourcing with Active Learning in Life Events Classification}
% If the paper title is too long for the running head, you can set
% an abbreviated paper title here
%
%
\author{Yunpeng Zhao\inst{1} \and
Mattia Prosperi\inst{1} \and
Tianchen Lyu\inst{1} \and
Yi Guo\inst{1} \and
Le Zhou\inst{2} \and
Jiang Bian\inst{1}
}
\authorrunning{Y. Zhao et al.}
% First names are abbreviated in the running head.
% If there are more than two authors, 'et al.' is used.
%
\institute{University of Florida, Gainesville FL 32611, USA\\
\email{\{yup111,m.prosperi,paint.94,yiguo,bianjiang\}@ufl.edu}  \and
University of Minnesota, Twin Cities, MN 55455, USA\\
\email{zhoul@umn.edu}\\
}
\maketitle              % typeset the header of the contribution
\begin{abstract}
Social media, especially Twitter, is being increasingly used for research with predictive analytics.  In social media studies, natural language processing (NLP) techniques are used in conjunction with expert-based, manual and qualitative analyses.  However, social media data are unstructured and must undergo complex manipulation for research use.  The manual annotation is the most resource and time-consum-ing process that multiple expert raters have to reach consensus on every item, but is essential to create gold-standard datasets for training NLP-based machine learning classifiers.  To reduce the burden of the manual annotation, yet maintaining its reliability, we devised a crowdsourcing pipeline combined with active learning strategies.  We demonstrated its effectiveness through a case study that identifies job loss events from individual tweets.  We used Amazon Mechanical Turk platform to recruit annotators from the Internet and designed a number of quality control measures to assure annotation accuracy.  We evaluated 4 different active learning strategies (i.e., least confident, entropy, vote entropy, and Kullback-Leibler divergence).  The active learning strategies aim at reducing the number of tweets needed to reach a desired performance of automated classification.  Results show that crowdsourcing is useful to create high-quality annotations and active learning helps in reducing the number of required tweets, although there was no substantial difference among the strategies tested.

\keywords{Social media  \and Crowdsourcing \and Active learning.}
\end{abstract}
\section{Introduction}
Micro-blogging social media platforms have become very popular in recent years.  One of the most popular platforms is Twitter, which allows users to broadcast short texts (i.e., 140 characters initially, and 280 characters in a recent platform update) in real time with almost no restrictions on content.  Twitter is a source of people’s attitudes, opinions, and thoughts toward the things that happen in their daily life.  Twitter data are publicly accessible through Twitter application programming interface (API); and there are several tools to download and process these data.  Twitter is being increasingly used as a valuable instrument for surveillance research and predictive analytics in many fields including epidemiology, psychology, and social sciences.  For example, Bian et al. explored the relation between promotional information and laypeople’s discussion on Twitter by using topic modeling and sentiment analysis \cite{bian_using_2017}.  Zhao et al. assessed the mental health signals among sexual and gender minorities using Twitter data  \cite{zhao_assessing_2018}.  Twitter data can be used to study and predict population-level targets, such as disease incidence \cite{eichstaedt_psychological_2015}, political trends \cite{gayo-avello_meta-analysis_2013}, earthquake detection \cite{sakaki_earthquake_2010}, and crime perdition \cite{hutchison_automatic_2012}, and individual-level outcomes or life events, such as job loss \cite{lossio-ventura_operational_2019}, depression \cite{leis_detecting_2019}, and adverse events \cite{wang_adverse_2018}.  Since tweets are unstructured textual data, natural language processing (NLP) and machine learning, especially deep learning nowadays, are often used for preprocessing and analytics.  However, for many studies\cite{finin_annotating_2010, mozetic_multilingual_2016, stowe_developing_2018}, especially those that analyze individual-level targets, manual annotations of several thousands of tweets, often by experts, is needed to create gold-standard training datasets, to be fed to the NLP and machine learning tools for subsequent, reliable automated processing of millions of tweets.  Manual annotation is obviously labor intense and time consuming. 

Crowdsourcing can scale up manual labor by distributing tasks to a large set of workers working in parallel instead of a single people working serially \cite{carenini_extractive_2008}.  Commercial platforms such as Amazon’s Mechanical Turk (MTurk, https://www.\\mturk.com/), make it easy to recruit a large crowd of people working remotely to perform time consuming manual tasks such as entity resolution \cite{arasu_active_2010, bellare_active_2012}, image or sentiment annotation \cite{vijayanarasimhan_cost-sensitive_2011, pak_twitter_2010}.  The annotation tasks published on MTurk can be done on a piecework basis and, given the very large pool of workers usually available (even by selecting a subset of those who have, say, a college degree), the tasks can be done almost immediately.  However, any crowdsourcing service that solely relies on human workers will eventually be expensive when large datasets are needed, that is often the case when creating training datasets for NLP and deep learning tasks.  Therefore, reducing the training dataset size (without losing performance and quality) would also improve efficiency while contain costs.

Query optimization techniques (e.g., active learning) can reduce the number of tweets that need to be labeled, while yielding comparable performance for the downstream machine learning tasks \cite{marcus_counting_2012,franklin_crowddb:_2011,parameswaran_crowdscreen:_2012}.  Active learning algorithms have been widely applied in various areas including NLP \cite{tang_active_2002} and image processing \cite{wang_cost-effective_2017}.  In a pool-based active learning scenario, data samples for training a machine learning algorithm (e.g., a classifier for identifying job loss events) are drawn from a pool of unlabeled data according to some forms of informativeness measure (a.k.a. active learning strategies \cite{settles_active_2009}), and then the most informative instances are selected to be annotated.  For a classification task, in essence, an active learning strategy should be able to pick the “best” samples to be labelled that will improve the classification performance the most.

In this study, we integrated active learning into a crowdsourcing pipeline for the classification of life events based on individual tweets.  We analyzed the quality of crowdsourcing annotations and then experimented with different machine/deep learning classifiers combined with different active learning strategies to answer the following two research questions (RQs): 
\begin{itemize}
  \item[$\bullet$] \textbf{RQ1.} How does (1) the amount of time that a human worker spends on and (2) the number of workers assigned to each annotation task impact the quality of an-notation results?
  \item[$\bullet$] \textbf{RQ2.} Which active learning strategy is most efficient and cost-effective to build event classification models using Twitter data?
\end{itemize}

\setlength{\belowcaptionskip}{-5pt}
\section{Methods}
\begin{figure}
\centering
\includegraphics[width=110mm,scale=1]{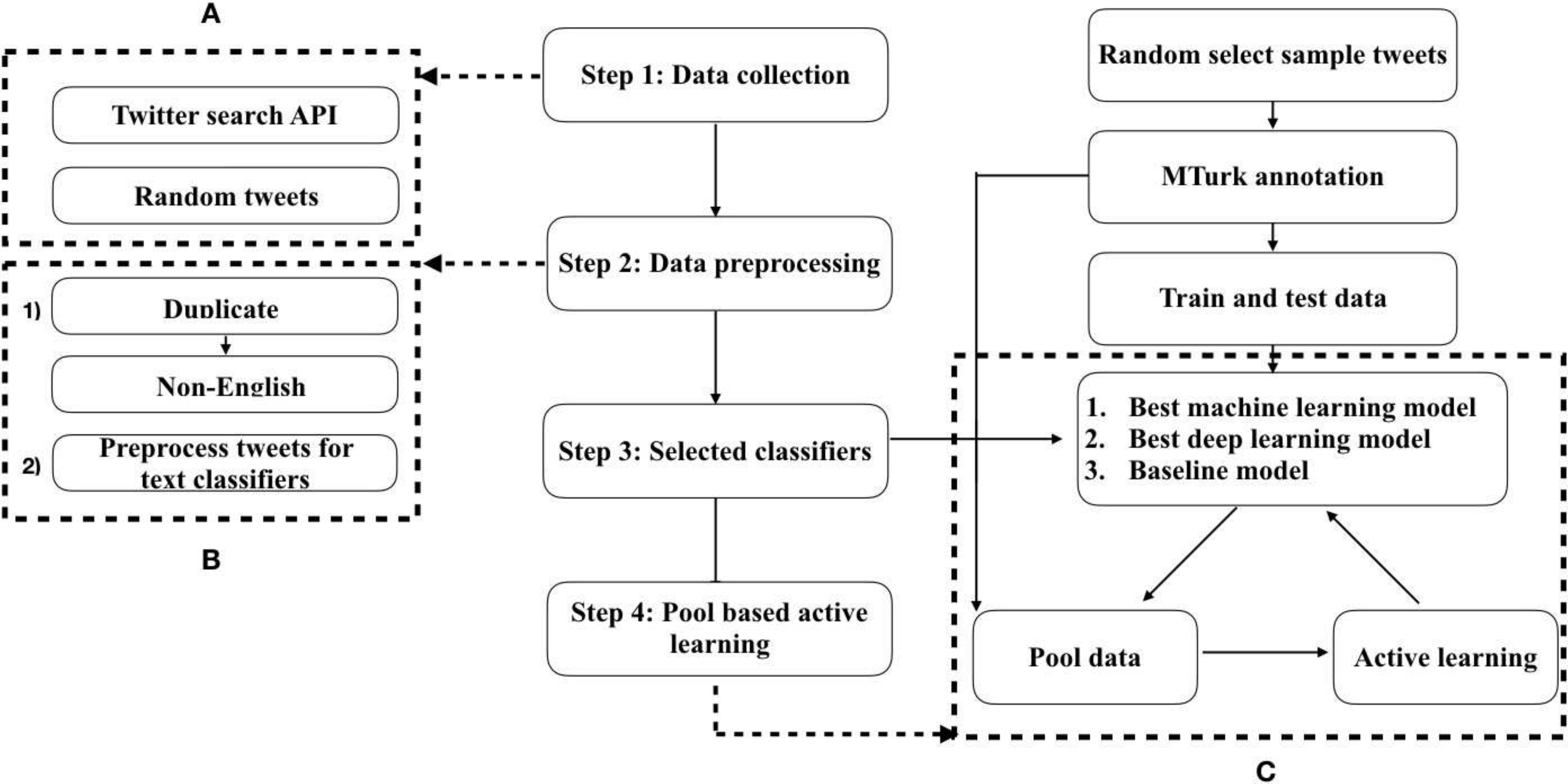}
\caption{The workflow of our Twitter analysis pipeline.} \label{fig1}
\end{figure}
We first collected tweets based on a list of job loss-related keywords.  We then randomly selected a set of sample tweets and had these tweets annotated (i.e., whether the tweet is a job loss event) using the Amazon MTurk platform.  With these annotated tweets, we then evaluated 4 different active learning strategies (i.e., least confi-dent, entropy, vote entropy, and Kullback-Leibler (KL) divergence) through simulations.

\subsection{Data Collection}
Our data were collected from two data sources based on a list of job loss-related keywords.  The keywords were developed using a snowball sampling process, where we started with an initial list of 8 keywords that indicates a job-loss event (e.g., “got fired” and “lost my job”).  Using these keywords, we then queried (1) Twitter’s own search engine (i.e., https://twitter.com/search-home?lang=en), and (2) a database of public random tweets that we have collected using the Twitter steaming application programming interface (API) from January 1, 2013 to December 30, 2017, to identify job loss-related tweets.  We then manually reviewed a sample of randomly selected tweets to discover new job loss-related keywords.  We repeated the search then review process iteratively until no new keywords were found.  Through this process, we found 33 keywords from the historical random tweet database and 57 keywords through Twitter web search.  We then (1) not only collected tweets based on the over-all of 68 unique keywords from the historical random tweet database, but also (2) crawled new Twitter data using Twitter search API from December 10, 2018 to December 26, 2018 (17 days).

% Please add the following required packages to your document preamble:
% \usepackage{multirow}

% Please add the following required packages to your document preamble:
% \usepackage{multirow}
% \usepackage{graphicx}

\subsection{Data Preprocessing}
We preprocessed the collected data to eliminate tweets that were (1) duplicated or (2) not written in English.  For building classifiers, we preprocessed the tweets following the preprocessing steps used by GloVe \cite{pennington_glove:_2014} with minor modifications as follows: (1) all hashtags (e.g., “\#gotfired”) were replaced with “$<$hashtag$>$ PHRASE” (e.g.,, “$<$hashtag$>$ gotfired”); (2) user mentions (e.g., “$@$Rob\_Bradley”) were replaced with “$<$user$>$”;  (3) web links (eg, “https://t.co/\\fMmFWAHEuM”) were replaced with “$<$url$>$”; and (4) all emojis were replaced with “$<$emoji$>$.”

\subsection{Classifier Selection}
Machine learning and deep learning have been wildly used in classification of tweets tasks.  We evaluated 8 different classifiers: 4 traditional machine learning models (i.e., logistic regress [LR], Naïve Bayes [NB], random forest [RF], and support vector machine [SVM]) and 4 deep learning models (i.e., convolutional neural network [CNN], recurrent neural network [RNN], long short-term memory [LSTM] RNN, and gated recurrent unit [GRU] RNN).  3,000 tweets out of 7,220 Amazon MTurk annotated dataset was used for classifier training (n = 2,000) and testing (n = 1,000).  The rest of MTurk annotated dataset were used for the subsequent active learning experiments.  Each classifier was trained 10 times and 95 confidence intervals (CI) for mean value were reported.  We explored two language models as the features for the classifiers (i.e., n-gram and word-embedding).  All the machine learning classifiers were developed with n-gram features; while we used both n-gram and word-embedding features on the CNN classifier to test which feature set is more suitable for deep learning classifiers.  CNN classifier with word embedding features had a better performance which is consistent with other studies \cite{le_comparative_2018, badjatiya_deep_2017}  We then selected one machine learning and one deep learning classifiers based on the prediction performance (i.e., F-score).  Logistic regression was used as the baseline classifier.

\subsection{Pool-based Active Learning}
In pool-based sampling for active learning, instances are drawn from a pool of samples according to some sort of informativeness measure, and then the most informative instances are selected to be annotated.  This is the most common scenario in active learning studies \cite{min_efficient_2017}.  The informativeness measures of the pool instances are called active learning strategies (or query strategies).  We evaluated 4 active learning strategies (i.e., least confident, entropy, vote entropy and KL divergence).   \textbf{Fig 1.C} shows the workflow of our pool-based active learning experiments: for a given active learning strategy and classifiers trained with an initial set of training data (1) the classifiers make predictions of the remaining to-be-labelled dataset; (2) a set of samples is selected using the specific active learning strategy and annotated by human reviewers; (3) the classifiers are retrained with the newly annotated set of tweets.  We repeated this process iteratively until the pool of data exhausts.  For the least confident and entropy active learning strategies, we used the best performed machine learn-ing classifier and the best performed deep learning classifier plus the baseline classifier (LR).  Note that vote entropy and KL divergence are query-by-committee strategies, which were tested upon three deep learning classifiers (i.e., CNN, RNN and LSTM) and three machine learning classifiers (i.e., LR, RF, and SVM) as two separate committees, respectively.

\section{Results}
\subsection{Data Collection}
Our data came from two different sources as shown in \textbf{Table 1}.  First, we collected 2,803,164 tweets using the Twitter search API \cite{noauthor_twitter_nodate} from December 10, 2018 to December 26, 2018 base on a list of job loss-related keywords (n = 68).  After filtering out duplicates and non-English tweets, 1,952,079 tweets were left.  Second, we used the same list of keywords to identify relevant tweets from a database of historical random public tweets we collected from January 1, 2013 to December 30, 2017.  We found 1,733,905 relevant tweets from this database.  Due to the different mechanisms behind the two Twitter APIs (i.e., streaming API vs. search API), the volumes of the tweets from the two data sources were significantly different.  For the Twitter search API, users can retrieve most of the public tweets related to the provided keywords within 10 to 14 days before the time of data collection; while the Twitter streaming API returns a random sample (i.e., roughly 1\% to 20\% varying across the years) of all public tweets at the time and covers a wide range of topics.  After integrating the tweets from the two data sources, there were 3,685,984 unique tweets.

\begin{table}[]\fontsize{8pt}{9pt}\selectfont
\caption{Descriptive statistics of job loss-related tweets from our two data sources.}\label{tab1}
\begin{tabular}{|l|l|l|l|}
\hline
\textbf{Data source}                                                                                & \textbf{Year} & \textbf{\# of tweets} & \textbf{\# of English tweets} \\ \hline
\multirow{5}{*}{\begin{tabular}[c]{@{}l@{}}Historical random public \\ tweet database\end{tabular}} & 2013          & 434,624               & 293,664                       \\ \cline{2-4} 
                                                                                                    & 2014          & 468,432               & 279,876                       \\ \cline{2-4} 
                                                                                                    & 2015          & 401,861               & 228,301                       \\ \cline{2-4} 
                                                                                                    & 2016          & 591,948               & 322,459                       \\ \cline{2-4} 
                                                                                                    & 2017          & 1,299,006             & 609,605                       \\ \hline
\begin{tabular}[c]{@{}l@{}}Collected through Twitter \\ search API\end{tabular}                     & 2018          & 2,803,164             & 1,952,079                     \\ \hline
                                                                                                    & Total         & 5,999,035             & 3,685,984                     \\ \hline
\end{tabular}%
\end{table}

\subsection{RQ1. How does (1) the amount of time that a human worker spends on and (2) the number of workers assigned to each annotation task impact the quality of annotation results?}
We randomly selected 7,220 tweets from our Twitter data based on keyword distributions and had those tweets annotated using workers recruited through Amazon MTurk.  Each tweet was also annotated by an expert annotator (i.e., one of the authors).  We treated the consensus answer of the crowdsourcing workers (i.e., at least 5 annotators for each tweet assignment) and the expert annotator as the gold-standard.  Using control tweets is a common strategy to identify workers who cheat (e.g., randomly select an answer without reading the instructions and/or tweets) on annotation tasks.  We introduced two control tweets in each annotation assignment, where each annotation assignment contains a total of 12 tweets (including the 2 control tweets).  Only responses with the two control tweets answered corrected were considered valid responses and the worker would receive the 10 cents incentive.  

The amount of time that a worker spends on a task is another factor associated with annotation quality.  We measured the time that one spent on clicking through the annotation task without thinking about the content and repeated the experiment five times.  The mean amount time spent on the task is 57.01 (95\% CI [47.19, 66.43]) seconds.  Thus, responses with less than 47 seconds were considered invalid regardless how the control tweets were answered. 
 
We then did two experiments to explore the relation between the amount of time that workers spend on annotation tasks and annotation quality.   \textbf{Fig 2. A.} shows annotation quality by selecting different amounts of lower cut-off time (i.e., only considering assignments where workers spent more time than the cut-off time as valid responses), which tests whether the annotation is of low quality when workers spent more time on the task.  The performance of the crowdsourcing workers was measured by the agreement (i.e., Cohan’s kappa) between labels from each crowdsourcing worker and the gold-standard labels.   \textbf{Fig 2. B.} shows annotation quality by selecting different upper cut-off time (i.e., keep assignments whose time consumption were less than the cut-off time), which tests whether the annotation is of low quality when workers spent less time on the task.  As shown in Fig. 2. A and B, it does not affect the annotation quality when a worker spent more time on the task; while, the annota-ion quality is significantly lower if the worker spent less than 90 seconds on the task.

\begin{figure}
\centering
\includegraphics[width=110mm,scale=1]{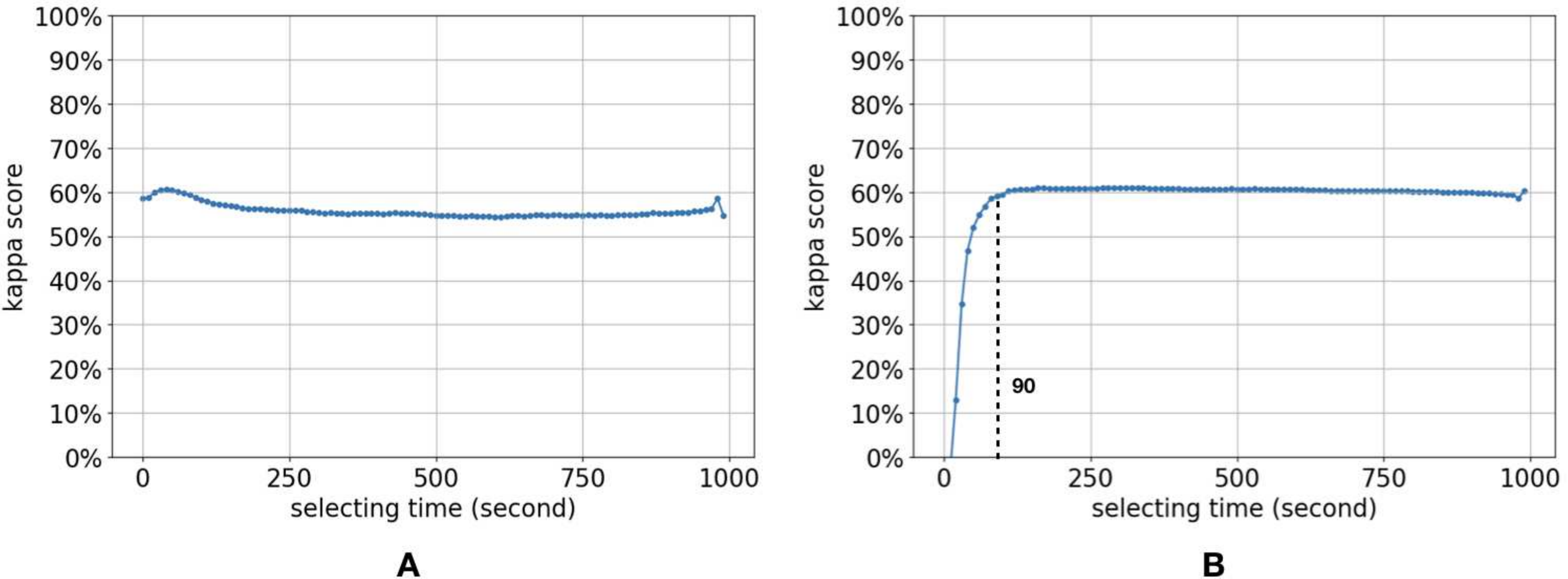}
\caption{Annotation quality by selecting different cut-off time.} \label{fig2}
\end{figure}

We also tested the annotation reliability (i.e., Fleiss’ Kappa score) between using 3 workers vs. using 5 workers.  The Fleiss’ kappa score of 3 workers is 0.53 (95\% CI [0.46, 0.61].  The Fleiss’ kappa score of 5 workers is 0.56 (95\% CI [0.51, 0.61].  Thus, using 3 workers vs. 5 workers does not make any difference on the annotation reliability, while it is obviously cheaper to use only 3 workers.  

\subsection{RQ2. Which active learning strategy is most efficient and cost-effective to build event classification models using Twitter data?  }
We randomly selected 3,000 tweets from the 7,220 MTurk annotated dataset to build the initial classifiers.  Two thousands out of 3,000 tweets were used to train the clas-sifiers and the rest 1,000 tweets were used as independent test dataset to benchmark their performance.  We explored 4 machine learning classifiers (i.e., Logistic Regression [LR], Naïve Bayes [NB], Random Forest [RF], and Support Vector Machine [SVM]) and 4 deep learning classifiers (i.e., Convolutional Neural Network [CNN], Recurrent Neural Network [RNN], Long Short-Term Memory [LSTM], and Gated Recurrent Unit [GRU]).  Each classifier was trained 10 times.  The performance was measured in terms of precision, recall, and F-score.  95\% confidence intervals (CIs) of the mean F-score across the ten runs were also reported.   \textbf{Table 2} shows the perfor-mance of classifiers.  We chose logistic regression as the baseline model.  RF and CNN were chosen for subsequent active learning experiments, since they outperformed other machine learning and deep learning classifiers.

% Please add the following required packages to your document preamble:
% \usepackage{multirow}
\begin{table}[]\fontsize{8pt}{9pt}\selectfont
\caption{The performance of machine learning and deep learning classifiers.}\label{tab2}
\begin{tabular}{|c|l|l|l|l|l|}
\hline
\multicolumn{1}{|l|}{\textbf{Feature}}                                                           & \textbf{Model name} & \textbf{Precision} & \textbf{Recall} & \textbf{F-score} & \textbf{95\% CIs of F-score} \\ \hline
\multirow{8}{*}{N-gram}                                                                          & \multicolumn{5}{l|}{Baseline}                                                                                \\ \cline{2-6} 
                                                                                                 & \textbf{LR}         & \textbf{0.75}      & \textbf{0.75}   & \textbf{0.74}    & \textbf{(0.74, 0.75)}        \\ \cline{2-6} 
                                                                                                 & \multicolumn{5}{l|}{Machine learning}                                                                        \\ \cline{2-6} 
                                                                                                 & NB                  & 0.73               & 0.73            & 0.72             & (0.72, 0.73)                 \\ \cline{2-6} 
                                                                                                 & RF                  & 0.74               & 0.74            & 0.74             & (0.74, 0.75)                 \\ \cline{2-6} 
                                                                                                 & SVM                 & 0.73               & 0.73            & 0.72             & (0.71, 0.73)                 \\ \cline{2-6} 
                                                                                                 & \multicolumn{5}{l|}{Deep learning}                                                                           \\ \cline{2-6} 
                                                                                                 & CNN                 & 0.71               & 0.71            & 0.72             & (0.71, 0.73)                 \\ \hline
\multicolumn{1}{|l|}{\multirow{4}{*}{\begin{tabular}[c]{@{}l@{}}Word-\\ embedding\end{tabular}}} & \textbf{CNN}        & \textbf{0.79}      & \textbf{0.79}   & \textbf{0.79}    & \textbf{(0.77, 0.80)}        \\ \cline{2-6} 
\multicolumn{1}{|l|}{}                                                                           & RNN                 & 0.75               & 0.75            & 0.74             & (0.74, 0.75)                 \\ \cline{2-6} 
\multicolumn{1}{|l|}{}                                                                           & LSTM                & 0.72               & 0.72            & 0.71             & (0.69, 0.72)                 \\ \cline{2-6} 
\multicolumn{1}{|l|}{}                                                                           & LSTM (GRU)          & 0.75               & 0.75            & 0.74             & (0.72, 0.76)                 \\ \hline
\end{tabular}
\end{table}

We implemented a pool-based active learning pipeline to test which classifier and active learning strategy is most efficient to build up an event classification classifier of Twitter data.  We queried the top 300 most “informative” tweets from the rest of the pool (i.e., excluding the tweets used for training the classifiers) at each iteration.   \textbf{Table 3} shows the active learning and classifier combinations that we evaluated.  The performance of the classifiers was measured by F-score.   \textbf{Fig 3} shows the results of the different active learning strategies combined with LR (i.e., the baseline), RF (i.e., the best performed machine learning model), and CNN (i.e., the best performed deep learning model).  For both machine learning models (i.e., LR and RF), using the entropy strategy can reach the optimal performance the quickest (i.e., the least amount of tweets).  While, the least confident algorithm does not have any clear advantages compared with random selection.  For deep learning model (i.e., CNN), none of the active learning strategies tested are useful to improve the CNN classifier’s performance.   \textbf{Fig 4} shows the results of query-by-committee algorithms (i.e., vote entropy and KL divergence) combined with machine learning and deep learning ensemble classifiers.  Query-by-committee algorithms are slightly better than random selection when it applied to machine learning ensemble classifier.  However, query-by-committee algorithms are not useful for the deep learning ensemble classifier.

\setcounter{footnote}{0}
\begin{table}[] \fontsize{8pt}{9pt}\selectfont
\caption{The active learning strategy and classifier combinations tested.}\label{tab3}
\resizebox{\textwidth}{!} {
\begin{tabular}{|l|l|}
\hline
\textbf{Model}                                                             & \textbf{Query strategies}                                                                         \\ \hline
LR                                                                         & Random query, least confident, entropy                                                            \\ \hline
RF                                                                         & Random query, least confident, entropy                                                            \\ \hline
CNN                                                                        & Random query, least confident, entropy                                                            \\ \hline
Ensemble\footnote                                                              & Vote entropy, KL divergence                                                                       \\ \hline
\multicolumn{2}{|l|}{\begin{tabular}[c]{@{}l@{}} {\setcounter{footnote}{0}\tiny\footnote amachine learning ensemble classifier: LR, RF, and SVM.,Deep learning ensemble classifier: CNN, RNN, and LSTM}\end{tabular}} \\ \hline
\end{tabular}}
\end{table}
\begin{figure}[hbt!]
\centering
\includegraphics[width=110mm,scale=0.9]{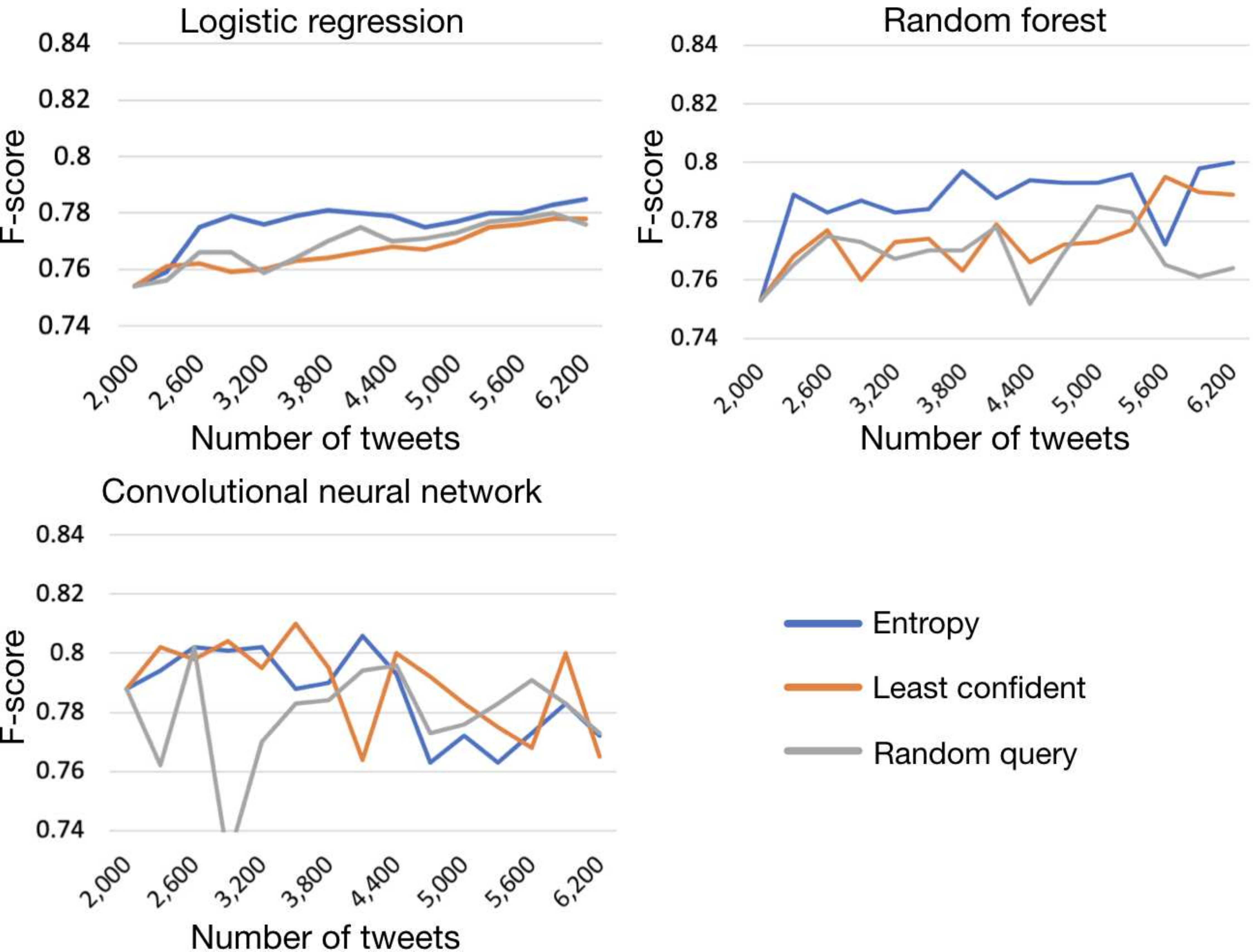}
\caption{The performance of active learning strategies combined with linear regression, random forest, and convolutional neural network.} \label{fig4}
\end{figure}
\begin{figure}[hbt!]
\centering
\includegraphics[width=110mm,scale=0.5]{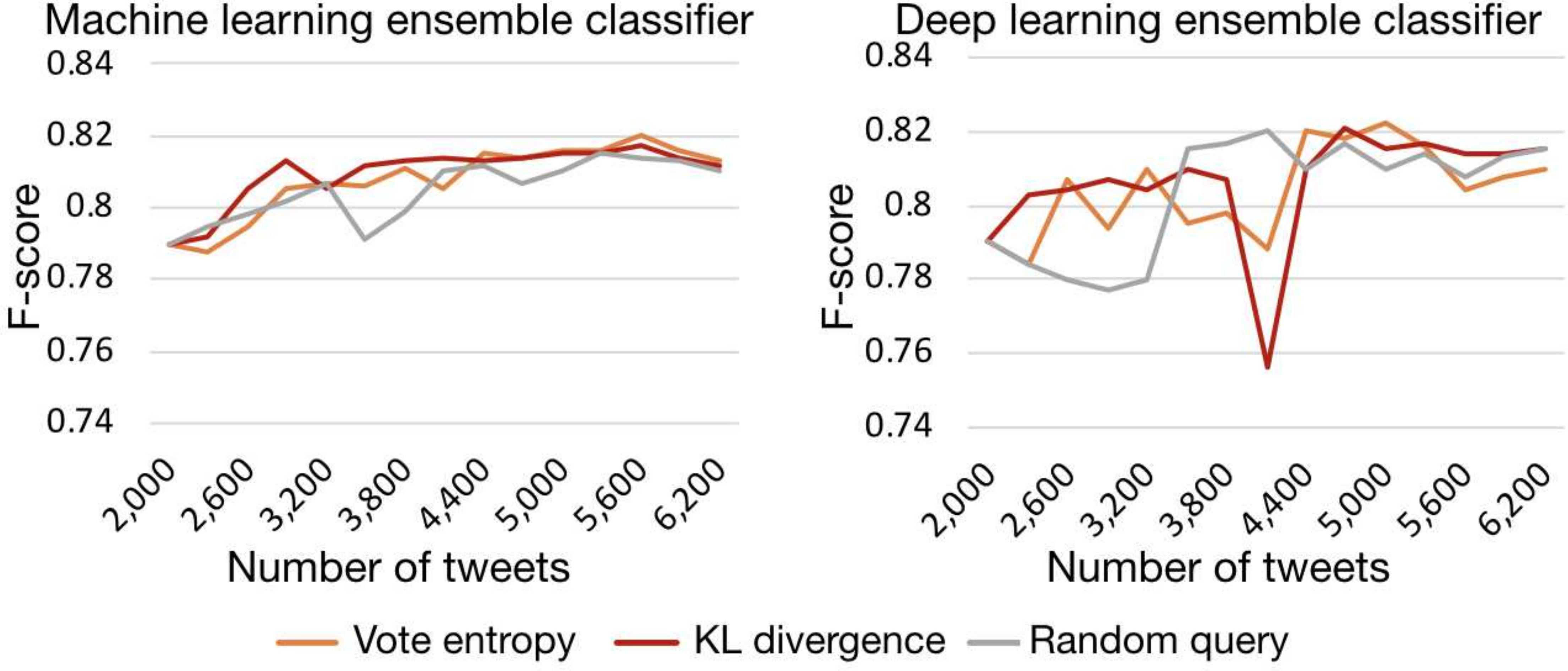}
\caption{The performance of active learning strategies combined with machine learning and deep learning ensemble classifiers.} \label{fig4}
\end{figure}

\section{Discussion}
The goal of our study was to test the feasibility of building classifiers by using crowdsourcing and active learning strategies.  We had 7,220 sample job loss-related tweets annotated using Amazon MTurk, tested 8 classification models, and evaluated 4 active learning strategies to answer our two RQs.

The key benefit of crowdsourcing is to have a large number of workers available to carry out tasks on a piecework basis.  This means that it is likely to get the crowd to start work on tasks almost immediately and be able to have a large number of tasks completed quickly.  However, even welltrained workers are only human and can make mistakes.  Our first RQ was to find an optimal and economical way to get reliable annotations from crowdsourcing.  Beyond using control tweets, we tested different cut-off time to assess how the amount of time workers spent on the task would affect annotation quality.  We found that the annotation quality is low if the tasks were finished within 90 seconds.  We also found that the annotation quality is not affected by the number of workers (i.e., between 3 worker group vs 5 worker group), which was also demonstrated by Mozafari et al \cite{mozafari_scaling_2014}.

In second RQ, we aimed to find which active learning strategy is most efficient and cost-effective to build event classification models using Twitter data.  We started with selecting representative machine learning and deep learning classifiers.  Among the 4 machine learning classifiers (i.e., LR, NB, RF, and SVM), LR and RF classifiers have the best performance on the task of identifying job loss events from tweets.  Among the 4 deep learning methods (i.e., CNN, RNN, LSTM, LSTM with GRU), CNN has the best performance. 

In active learning, the learning algorithm is set to proactively select a subset of available examples to be manually labeled next from a pool of yet unlabeled instances.  The fundamental idea behind the concept is that a machine learning algorithm could potentially achieve a better accuracy quicker and using fewer training data if it were allowed to choose the most informative data it wants to learn from.  In our experiment, we found that the entropy algorithm is the best way to build machine learning models fast and efficiently.  Vote entropy and KL divergence, the query-by-committee active learning methods are helpful for the training of machine learning ensemble classifiers.  However, all the active learning strategies we tested do not work well with deep learning model (i.e., CNN) or deep learning-based ensemble classifier.

We also recognize the limitations of our study.  First, we only tested 5 classifiers (i.e., LR, RF, CNN, a machine learning ensemble classifier, and a deep learning classifier) and 4 active learning strategies (i.e., least confident, entropy, vote entropy, KL divergence).  Other state-of-art methods for building tweet classifiers (e.g., BERT \cite{devlin_bert:_2018}) and other active learning strategies (e.g., variance reduction \cite{yang_variance_2018}) are worth exploring.  Second, other crowdsourcing quality control methods such as using prequalification questions to identify high-quality workers also warrant further investigations.  Third, the crowdsourcing and active learning pipeline can potentially be applied to other data and tasks.  However, more experiments are needed to test the fea-sibility.  Fourth, the current study only focused on which active learning strategy is most efficient and cost-effective to build event classification models using crowdsourcing labels.  Other research questions such as how the correctness of the crowdsourced labels would impact classifier performance warrant future investigations.

In sum, our study demonstrated that crowdsourcing with active learning is a possible way to build up machine learning classifiers efficiently.  However, active learning strategies do not benefit deep learning classifiers in our study.

\section*{Acknowledgement}
This study was supported by NSF Award \#1734134.
% the environments 'definition', 'lemma', 'proposition', 'corollary',
% 'remark', and 'example' are defined in the LLNCS documentclass as well.
%
%
% ---- Bibliography ----
%
% BibTeX users should specify bibliography style 'splncs04'.
% References will then be sorted and formatted in the correct style.
%
\bibliographystyle{splncs03_unsrt}
 \bibliography{twitter_jobloss_journal_new}

\begin{thebibliography}{10}
\providecommand{\url}[1]{\texttt{#1}}
\providecommand{\urlprefix}{URL }

\bibitem{bian_using_2017}
Bian, J., Zhao, Y., Salloum, R.G., Guo, Y., Wang, M., Prosperi, M., Zhang, H.,
  Du, X., Ramirez-Diaz, L.J., He, Z., Sun, Y.: Using {Social} {Media} {Data} to
  {Understand} the {Impact} of {Promotional} {Information} on {Laypeople}’s
  {Discussions}: {A} {Case} {Study} of {Lynch} {Syndrome}. Journal of Medical
  Internet Research  19(12),  e414 (2017)

\bibitem{zhao_assessing_2018}
Zhao, Y., Guo, Y., He, X., Huo, J., Wu, Y., Yang, X., Bian, J.: Assessing
  {Mental} {Health} {Signals} {Among} {Sexual} and {Gender} {Minorities} using
  {Twitter} {Data}. In: 2018 {IEEE} {International} {Conference} on
  {Healthcare} {Informatics} {Workshop} ({ICHI}-{W}). pp. 51--52. IEEE, New
  York, NY (2018)

\bibitem{eichstaedt_psychological_2015}
Eichstaedt, J.C., Schwartz, H.A., Kern, M.L., Park, G., Labarthe, D.R.,
  Merchant, R.M., Jha, S., Agrawal, M., Dziurzynski, L.A., Sap, M., Weeg, C.,
  Larson, E.E., Ungar, L.H., Seligman, M.E.P.: Psychological {Language} on
  {Twitter} {Predicts} {County}-{Level} {Heart} {Disease} {Mortality}.
  Psychological Science  26(2),  159--169 (2015)

\bibitem{gayo-avello_meta-analysis_2013}
Gayo-Avello, D.: A {Meta}-{Analysis} of {State}-of-the-{Art} {Electoral}
  {Prediction} {From} {Twitter} {Data}. Social Science Computer Review  31(6),
  649--679 (2013)

\bibitem{sakaki_earthquake_2010}
Sakaki, T., Okazaki, M., Matsuo, Y.: Earthquake shakes {Twitter} users:
  real-time event detection by social sensors. In: Proceedings of the 19th
  international conference on {World} wide web - {WWW} '10. p. 851. ACM Press,
  Raleigh, North Carolina, USA (2010)

\bibitem{hutchison_automatic_2012}
Wang, X., Gerber, M.S., Brown, D.E.: Automatic {Crime} {Prediction} {Using}
  {Events} {Extracted} from {Twitter} {Posts}. In: Hutchison, D., Kanade, T.,
  Kittler, J., Kleinberg, J.M., Mattern, F., Mitchell, J.C., Naor, M.,
  Nierstrasz, O., Pandu~Rangan, C., Steffen, B., Sudan, M., Terzopoulos, D.,
  Tygar, D., Vardi, M.Y., Weikum, G., Yang, S.J., Greenberg, A.M., Endsley, M.
  (eds.) Social {Computing}, {Behavioral} - {Cultural} {Modeling} and
  {Prediction}, vol. 7227, pp. 231--238. Springer Berlin Heidelberg, Berlin,
  Heidelberg (2012)

\bibitem{lossio-ventura_operational_2019}
Du, X., Bian, J., Prosperi, M.: An {Operational} {Deep} {Learning} {Pipeline}
  for {Classifying} {Life} {Events} from {Individual} {Tweets}. In:
  Lossio-Ventura, J.A., Muñante, D., Alatrista-Salas, H. (eds.) Information
  {Management} and {Big} {Data}, vol. 898, pp. 54--66. Springer International
  Publishing, Cham (2019)

\bibitem{leis_detecting_2019}
Detecting {Signs} of {Depression} in {Tweets} in {Spanish}: {Behavioral} and
  {Linguistic} {Analysis}. Journal of Medical Internet Research  21(6),  e14199
  (2019)

\bibitem{wang_adverse_2018}
Wang, J., Zhao, L., Ye, Y., Zhang, Y.: Adverse event detection by integrating
  twitter data and {VAERS}. Journal of Biomedical Semantics  9(1) (2018)

\bibitem{finin_annotating_2010}
Finin, T., Murnane, W., Karandikar, A., Keller, N., Martineau, J., Dredze, M.:
  Annotating {Named} {Entities} in {Twitter} {Data} with {Crowdsourcing}. In:
  Proceedings of the {NAACL} {HLT} 2010 {Workshop} on {Creating} {Speech} and
  {Language} {Data} with {Amazon}'s {Mechanical} {Turk}. pp. 80--88.
  Association for Computational Linguistics, Los Angeles (2010)

\bibitem{mozetic_multilingual_2016}
Mozetič, I., Grčar, M., Smailović, J.: Multilingual {Twitter} {Sentiment}
  {Classification}: {The} {Role} of {Human} {Annotators}. PLOS ONE  11(5),
  e0155036 (2016)

\bibitem{stowe_developing_2018}
Stowe, K., Palmer, M., Anderson, J., Kogan, M., Palen, L., Anderson, K.M.,
  Morss, R., Demuth, J., Lazrus, H.: Developing and {Evaluating} {Annotation}
  {Procedures} for {Twitter} {Data} during {Hazard} {Events}. In: Proceedings
  of the {Joint} {Workshop} on {Linguistic} {Annotation}, {Multiword}
  {Expressions} and {Constructions} ({LAW}-{MWE}-{CxG}-2018). pp. 133--143.
  Association for Computational Linguistics, Santa Fe, New Mexico, USA (2018)

\bibitem{carenini_extractive_2008}
Carenini, G., Cheung, J.C.K.: Extractive vs. {NLG}-based {Abstractive}
  {Summarization} of {Evaluative} {Text}: {The} {Effect} of {Corpus}
  {Controversiality}. In: Proceedings of the {Fifth} {International} {Natural}
  {Language} {Generation} {Conference}. pp. 33--41. Association for
  Computational Linguistics, Salt Fork, Ohio, USA (2008)

\bibitem{arasu_active_2010}
Arasu, A., Götz, M., Kaushik, R.: On active learning of record matching
  packages. In: Proceedings of the 2010 international conference on
  {Management} of data - {SIGMOD} '10. p. 783. ACM Press, Indianapolis,
  Indiana, USA (2010)

\bibitem{bellare_active_2012}
Bellare, K., Iyengar, S., Parameswaran, A.G., Rastogi, V.: Active sampling for
  entity matching. In: Proceedings of the 18th {ACM} {SIGKDD} international
  conference on {Knowledge} discovery and data mining - {KDD} '12. p. 1131. ACM
  Press, Beijing, China (2012)

\bibitem{vijayanarasimhan_cost-sensitive_2011}
Vijayanarasimhan, S., Grauman, K.: Cost-{Sensitive} {Active} {Visual}
  {Category} {Learning}. International Journal of Computer Vision  91(1),
  24--44 (2011)

\bibitem{pak_twitter_2010}
Pak, A., Paroubek, P.: Twitter as a {Corpus} for {Sentiment} {Analysis} and
  {Opinion} {Mining}. In: {LREC} (2010)

\bibitem{marcus_counting_2012}
Marcus, A., Karger, D., Madden, S., Miller, R., Oh, S.: Counting with the
  crowd. Proceedings of the VLDB Endowment  6(2),  109--120 (2012)

\bibitem{franklin_crowddb:_2011}
Franklin, M.J., Kossmann, D., Kraska, T., Ramesh, S., Xin, R.: {CrowdDB}:
  answering queries with crowdsourcing. In: Proceedings of the 2011
  international conference on {Management} of data - {SIGMOD} '11. p.~61. ACM
  Press, Athens, Greece (2011)

\bibitem{parameswaran_crowdscreen:_2012}
Parameswaran, A.G., Garcia-Molina, H., Park, H., Polyzotis, N., Ramesh, A.,
  Widom, J.: {CrowdScreen}: algorithms for filtering data with humans. In:
  Proceedings of the 2012 international conference on {Management} of {Data} -
  {SIGMOD} '12. p. 361. ACM Press, Scottsdale, Arizona, USA (2012)

\bibitem{tang_active_2002}
Tang, M., Luo, X., Roukos, S.: Active {Learning} for {Statistical} {Natural}
  {Language} {Parsing}. In: Proceedings of the 40th {Annual} {Meeting} of the
  {Association} for {Computational} {Linguistics}. pp. 120--127. Association
  for Computational Linguistics, Philadelphia, Pennsylvania, USA (2002)

\bibitem{wang_cost-effective_2017}
Wang, K., Zhang, D., Li, Y., Zhang, R., Lin, L.: Cost-{Effective} {Active}
  {Learning} for {Deep} {Image} {Classification}. IEEE Transactions on Circuits
  and Systems for Video Technology  27(12),  2591--2600 (2017), arXiv:
  1701.03551

\bibitem{settles_active_2009}
Settles, B.: Active {Learning} {Literature} {Survey}. Computer {Sciences}
  {Technical} {Report} 1648, University of Wisconsin–Madison (2009)

\bibitem{pennington_glove:_2014}
Pennington, J., Socher, R., Manning, C.D.: {GloVe}: {Global} {Vectors} for
  {Word} {Representation}. In: Empirical {Methods} in {Natural} {Language}
  {Processing} ({EMNLP}). pp. 1532--1543 (2014)

\bibitem{le_comparative_2018}
Le, H.P., Le, A.C.: A {Comparative} {Study} of {Neural} {Network} {Models} for
  {Sentence} {Classification}. 2018 5th NAFOSTED Conference on Information and
  Computer Science (NICS) pp. 360--365 (2018)

\bibitem{badjatiya_deep_2017}
Badjatiya, P., Gupta, S., Gupta, M., Varma, V.: Deep {Learning} for {Hate}
  {Speech} {Detection} in {Tweets}. In: Proceedings of the 26th {International}
  {Conference} on {World} {Wide} {Web} {Companion} - {WWW} '17 {Companion}. pp.
  759--760. ACM Press, Perth, Australia (2017)

\bibitem{min_efficient_2017}
Min, X., Shi, Y., Cui, L., Yu, H., Miao, Y.: Efficient {Crowd}-{Powered}
  {Active} {Learning} for {Reliable} {Review} {Evaluation}. In: Proceedings of
  the 2nd {International} {Conference} on {Crowd} {Science} and {Engineering} -
  {ICCSE}'17. pp. 136--143. ACM Press, Beijing, China (2017)

\bibitem{noauthor_twitter_nodate}
Twitter, I.: Twitter developer {API} reference index (2020)

\bibitem{mozafari_scaling_2014}
Mozafari, B., Sarkar, P., Franklin, M., Jordan, M., Madden, S.: Scaling up
  crowd-sourcing to very large datasets: a case for active learning.
  Proceedings of the VLDB Endowment  8(2),  125--136 (2014)

\bibitem{devlin_bert:_2018}
Devlin, J., Chang, M.W., Lee, K., Toutanova, K.: {BERT}: {Pre}-training of
  {Deep} {Bidirectional} {Transformers} for {Language} {Understanding}. arXiv
  preprint arXiv:1810.04805  (2018)

\bibitem{yang_variance_2018}
Yang, Y., Loog, M.: A variance maximization criterion for active learning.
  Pattern Recognition  78,  358--370 (2018)

\end{thebibliography}

\end{document}